\documentclass[conference]{IEEEtran}
\IEEEoverridecommandlockouts
\renewcommand{\thesection}{\Roman{section}}
\usepackage{booktabs}
\usepackage{fancyhdr}
\usepackage{algorithm}
\usepackage[usenames, dvipsnames]{xcolor}
\usepackage{algorithmic}
\usepackage{colortbl}
\usepackage{caption}
\usepackage{graphicx}
\usepackage{array}
\usepackage{adjustbox}
\usepackage[colorlinks=true, linkcolor=black, urlcolor=blue, citecolor=blue]{hyperref}

\definecolor{myblue}{RGB}{10, 150, 200}

\usepackage{cite}
\usepackage[most]{tcolorbox}
\usepackage{multirow}
\usepackage{amsmath,amssymb,amsfonts}
\usepackage{graphicx}
\usepackage{textcomp}
\usepackage{placeins}  
\usepackage{stfloats}
\usepackage{float}
\usepackage{comment}

\definecolor{highlightColor}{HTML}{E6FFE6}

\usepackage{xcolor}
\def\BibTeX{{\rm B\kern-.05em{\sc i\kern-.025em b}\kern-.08em
    T\kern-.1667em\lower.7ex\hbox{E}\kern-.125emX}}
    
\fancypagestyle{firstpage}{
  \fancyhf{}
  \fancyhead[L]{\small 2025 International Conference on Quantum Photonics, Artificial Intelligence, and Networking (QPAIN) \\ 31 July -- 2 August 2025, Rangpur, Bangladesh}
  \fancyfoot[L]{\small Accepted in 1st QPAIN, 2025}

}

\title{From Chat to Checkup: Can Large Language Models Assist in Diabetes Prediction?}

\author{ 
    \IEEEauthorblockN{Shadman Sakib, Oishy Fatema Akhand, Ajwad Abrar\\}
    \IEEEauthorblockA{Department of Computer Science and Engineering, Islamic University of Technology, Gazipur, Bangladesh\\}
    \IEEEauthorblockA{Email: \{shadmansakib20, oishyfatema, ajwadabrar\}@iut-dhaka.edu}
}
\hypersetup{
    colorlinks=true,
    linkcolor=blue,    
    citecolor=blue,     
    urlcolor=blue      
}

\begin{document}
\maketitle
\thispagestyle{firstpage}
\begin{abstract}
While Machine Learning (ML) and Deep Learning (DL) models have been widely used for diabetes prediction, the use of Large Language Models (LLMs) for structured numerical data is still not well explored. In this study, we test the effectiveness of LLMs in predicting diabetes using zero-shot, one-shot, and three-shot prompting methods. We conduct an empirical analysis using the Pima Indian Diabetes Database (PIDD). We evaluate six LLMs, including four open-source models: Gemma-2-27B, Mistral-7B, Llama-3.1-8B, and Llama-3.2-2B. We also test two proprietary models: GPT-4o and Gemini Flash 2.0. In addition, we compare their performance with three traditional machine learning models: Random Forest, Logistic Regression, and Support Vector Machine (SVM). We use accuracy, precision, recall, and F1-score as evaluation metrics. Our results show that proprietary LLMs perform better than open-source ones, with GPT-4o and Gemma-2-27B achieving the highest accuracy in few-shot settings. Notably, Gemma-2-27B also outperforms the traditional ML models in terms of F1-score. However, there are still issues such as performance variation across prompting strategies and the need for domain-specific fine-tuning. This study shows that LLMs can be useful for medical prediction tasks and encourages future work on prompt engineering and hybrid approaches to improve healthcare predictions.
\end{abstract}

\begin{IEEEkeywords}
Large Language Models (LLMs), Medical Data Prediction, Prompt Engineering, Diabetes Prediction, Machine Learning in Healthcare
\end{IEEEkeywords}

\section{Introduction}
Diabetes Mellitus (DM) refers to a group of chronic metabolic disorders characterized by persistent hyperglycemia, \cite{goyal2023type} and can give rise to serious complications, including cardiovascular disease, kidney failure, and neuropathy \cite{american2014diagnosis}. Worldwide, this disorder affects approximately 537 million adults \cite{hossain2024diabetes}, and the rising incidence of diabetes around the globe has made early diagnosis and treatment vital to reducing its burden. In fact, the pooled prevalence of diabetes in South Asia has increased from 11.3\% (2000--04) to 22.3\% (2020--24) \cite{ranasinghe2024rising}. Numerous studies have used the Pima Indian Diabetes Database (PIDD) \cite{smith1988} to assess different Machine Learning (ML) and Deep Learning (DL) models for the prediction of diabetes \cite{naz2020deep, mujumdar2019diabetes, chang2023pima}. 

The results clearly indicate that while traditional models like Decision Trees (DT) and Naïve Bayes (NB) provide interpretable classifications, deep learning techniques such as Artificial Neural Networks (ANN) \cite{agatonovic2000basic} and Recurrent Neural Networks (RNN) \cite{sherstinsky2020fundamentals} attain remarkable accuracy, typically surpassing 90\% \cite{naz2020deep, mujumdar2019diabetes}. Techniques for feature selection, like Recursive Feature Elimination (RFE) in conjunction with Gated Recurrent Units (GRU), have been implemented to enhance predictive accuracy \cite{shams2025novel}. These approaches, however, constrain their application in practical healthcare environments due to their substantial computational resource requirements and extended training durations \cite{chang2023pima}.

LLMs have shown strong performance in the field of bioinformatics, often matching fine-tuned models like BanglaT5, despite the lack of task-specific fine-tuning \cite{abrar2025performanceevaluationlargelanguage}. Models like Mistral and GPT-based architectures excel in complex reasoning tasks without domain-specific training \cite{mistral_large_2024}. LLMs are widely adopted in natural language and multimodal tasks, and several studies have reported instances of bias in their outputs \cite{abrar2025religiousbiaslandscapelanguage}. However, their ability to generalize to structured tabular data through prompt-based approaches remains relatively underexplored. The key contributions of this study are as follows:

\begin{itemize}  
    \item We evaluate six LLMs for diabetes prediction using the PIDD dataset, and compare their performance with traditional ML models including Logistic Regression, Random Forest, and Support Vector Machine (SVM).  
    \item We analyze the impact of zero-shot, one-shot, and three-shot prompting strategies on the prediction accuracy of LLMs when applied to structured numerical medical data.  
    \item We identify the strengths and limitations of LLMs in medical prediction tasks and discuss their feasibility for real-world healthcare applications.  
\end{itemize}

This study explores the use of natural language prompts to represent structured medical data and evaluates the zero-shot, one-shot, and three-shot performance of both open-source and proprietary LLMs. To support the reliability of this evaluation, we include repeated runs and compare results with traditional ML classifiers trained on the same dataset. This work offers an early comparison between prompt-based LLM inference and conventional ML methods on the PIDD dataset, conducted without fine-tuning or model-specific adaptations.\footnote{Code and resources are available at: \url{https://github.com/ShadmanSakib44/Diabetes_Prediction}}

\begin{figure*}[t]
\centering
\includegraphics[width=\textwidth]{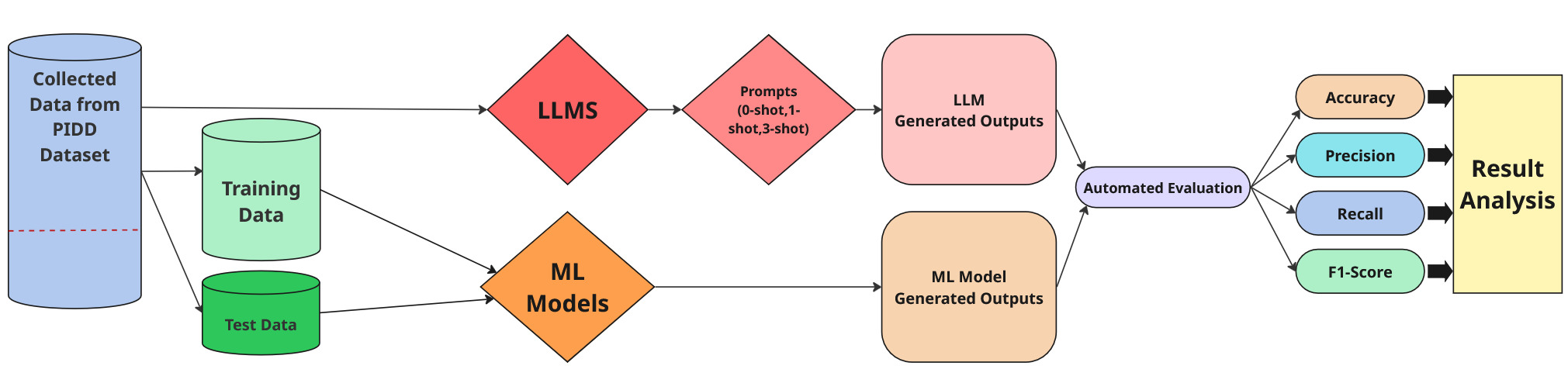}
\caption{Workflow of the experimental pipeline from data preparation to evaluation.}
\label{fig:workflow_diagram}
\end{figure*}
\section{Literature Review}
ML, DL, and their hybrid approaches have received significant attention in diabetes prediction. These methods have led to substantial progress across both traditional and advanced computational techniques \cite{mujumdar2019diabetes, naz2020deep, chang2023pima, shams2025novel}. However, the application of LLMs to structured numerical data, particularly in medical domains such as the Pima Indian Diabetes Dataset (PIDD), remains relatively underexplored. Despite the growing capabilities of LLMs, limited research has examined their effectiveness in this context, highlighting the need for further investigation—especially in low-resource medical settings \cite{naz2020deep}.

Prior research has shown the efficacy of machine learning and deep learning models in diabetes prediction tasks. Mujumdar et al. conducted a comparative study of multiple machine learning techniques utilizing the PIDD dataset alongside an additional dataset containing 800 records and 10 attributes \cite{mujumdar2019diabetes}.  Among all the supervised learning algorithms they used, their results indicated that Logistic Regression achieved the highest accuracy of 96\%, which was evaluated using metrics like classification accuracy, confusion matrix, and F1-score, emphasizing the importance of feature selection in improving model performance. Similarly, Chang et al. explored interpretable ML models for diabetes prediction in addition to proposing an e-diagnosis system. Their study indicated that Naïve Bayes performed well with fine-tuned feature selection with an accuracy of 77.83\% when reduced to 5 features, while Random Forest excelled with a larger feature set (79.57\% accuracy, 89.40\% precision, 86.24\% AUC) \cite{chang2023pima}.

Further research has investigated the efficacy of deep learning models for diabetes classification. Naz et al. systematically evaluated Artificial Neural Networks (ANN), deep learning, and traditional ML classifiers such as Decision Trees and Naïve Bayes \cite{naz2020deep}. Their findings indicated that DL models, particularly ANN, achieved the highest accuracy (98.07\%) on the PIDD dataset, while Decision Tree (DT) achieved 96.62\% accuracy, ANN achieved 90.34\% accuracy, and Naïve Bayes achieved 76.33\% accuracy. Deep learning approaches have been shown to be far superior in extracting complex data patterns.

Recent advancements have also explored hybrid models combining ML and DL techniques. Shams et al. proposed a Recursive Feature Elimination (RFE) with Gated Recurrent Units (GRU) model for diabetes prediction, achieving an accuracy of 90.70\% \cite{shams2025novel}. They integrated RFE for feature selection and GRU for handling feature dependencies, thus highlighting the effectiveness of feature selection in enhancing model performance.

Mousa et al. conducted an assessment of LSTM, Convolutional Neural Networks (CNN), and Random Forest (RF) models using the PIDD dataset. In this study, LSTM outperformed RF and CNN, achieving an accuracy of 85\% and AUC-ROC of 0.89 \cite{mousa2023comparative}. Meanwhile, Kalaiselvi et al. analyzed diabetes prediction using the Support Vector Machine (SVM) and K-Nearest Neighbors (KNN) in the PIDD dataset, where SVM showed 85.6\% accuracy and KNN 82.3\% accuracy, resulting in SVM having the best performance \cite{kalaiselvi2022analysis}. The highest accuracies reported in previous studies using the PIDD dataset are summarized in \autoref{table:best_accuracies}.

\begin{table}[htbp]
\centering
\caption{Comparison of Best Algorithms Achieved in Referenced Studies on Diabetes Prediction Using PIDD Dataset}
\label{table:best_accuracies}
\begin{adjustbox}{max width=\columnwidth}
\begin{tabular}{l c l}
    \toprule
    \textbf{Paper} & \textbf{Accuracy (\%)} & \textbf{Best Algorithms/Methods} \\
    \midrule
    Mujumdar et al. \cite{mujumdar2019diabetes} & 77 & Gradient Boost Classifier \\
    Mujumdar et al. \cite{mujumdar2019diabetes} & 77 & LDA \\
    Mujumdar et al. \cite{mujumdar2019diabetes} & 77 & AdaBoost Classifier \\
    Naz et al. \cite{naz2020deep} & 98.07 & Deep Learning \\
    Mousa et al. \cite{mousa2023comparative} & 85 & LSTM \\
    Chang et al. \cite{chang2023pima} & 79.57 & Random Forest (full dataset) \\
    Chang et al. \cite{chang2023pima} & 79.13 & Naïve Bayes (3-factor) \\
    Chang et al. \cite{chang2023pima} & 77.83 & Naïve Bayes (5-factor) \\
    Shams et al. \cite{shams2025novel} & 90.7 & Proposed RFE-GRU \\
    Kalaiselvi et al. \cite{kalaiselvi2022analysis} & 85.6 & SVM \\
    \bottomrule
\end{tabular}

\end{adjustbox}
\end{table}

Despite recent advancements, a comprehensive evaluation of LLMs for structured numerical data classification remains lacking. This study bridges this gap by assessing the zero-shot, one-shot, and three-shot prompting performance of both open-source and closed-source state-of-the-art LLMs for diabetes prediction using the PIDD dataset, providing valuable insights into their potential for real-world healthcare applications.

\section{Methodology}

This study evaluates the effectiveness of LLMs in predicting diabetes using zero-shot, one-shot, and three-shot settings, and compares these approaches with traditional machine learning models. To achieve this, we utilized the PIDD dataset\footnote{ \url{https://www.kaggle.com/datasets/uciml/pima-indians-diabetes-database}} and evaluated the performance of six LLMs for the assessment, alongside three traditional ML classifiers: Logistic Regression, Random Forest, and SVM. Each LLM and prompting setup was assessed across three separate runs, with the mean performance reported to enhance reliability and reduce the effect of random variation.

\subsection{Dataset}

The PIDD dataset, provided by the National Institute of Diabetes and Digestive and Kidney Diseases, is a widely used benchmark dataset in medical machine learning. It consists of health-related data from female patients of Pima Indian heritage, all of whom are at least 21 years old. The dataset is primarily used for binary classification tasks, aiming to predict whether a patient has diabetes based on clinical measurements.

The dataset includes 768 instances with 9 attributes, comprising 8 numeric predictor variables and 1 binary target variable.

\begin{table}[htbp]
\centering
\caption{Attributes of the Pima Indian Diabetes Dataset}
\label{table:pima_summary}
\begin{adjustbox}{max width=\linewidth}
\begin{tabular}{llc}
\toprule
\textbf{Feature} & \textbf{Description} & \textbf{Data Type} \\
\midrule
Pregnancies & Number of times pregnant & Integer \\
Glucose & Plasma glucose concentration (mg/dL) & Integer \\
BloodPressure & Diastolic blood pressure (mm Hg) & Integer \\
SkinThickness & Triceps skin fold thickness (mm) & Integer \\
Insulin & 2-hour serum insulin (mu U/ml) & Integer \\
BMI & Body mass index (kg/m\textsuperscript{2}) & Float \\
DiabetesPedigreeFunction & Diabetes risk based on family history & Float \\
Age & Age in years & Integer \\
Outcome & Diabetes diagnosis (0 = No, 1 = Yes) & Integer (Binary) \\
\bottomrule
\end{tabular}
\end{adjustbox}
\end{table}

The dataset exhibits a moderately imbalanced class distribution, with approximately 65.1\% labeled as non-diabetic (Outcome = 0) and 34.9\% as diabetic (Outcome = 1). The structured nature of the dataset and its widespread use in diabetes-related studies make it an ideal benchmark for assessing model performance on medical prediction tasks.

\subsection{Models}

Our evaluation encompasses a total of six LLMs, comprising two proprietary and four open-source models. Additionally, three traditional machine learning models—Logistic Regression, Random Forest, and SVM—were implemented using scikit-learn with default hyperparameters. For the ML models, features were standardized using z-score normalization, and the dataset was split into 80\% training and 20\% testing. This pipeline provides a classical benchmark against which LLM performance can be compared.

To provide a comprehensive analysis, we categorize LLMs based on their accessibility and licensing. In this section, we focus on detailing the free and open-source models, highlighting their characteristics and relevance to our study.

\begin{figure*}[t]
\vspace{-1cm} 
  \centering
  \includegraphics[width=.8\textwidth]{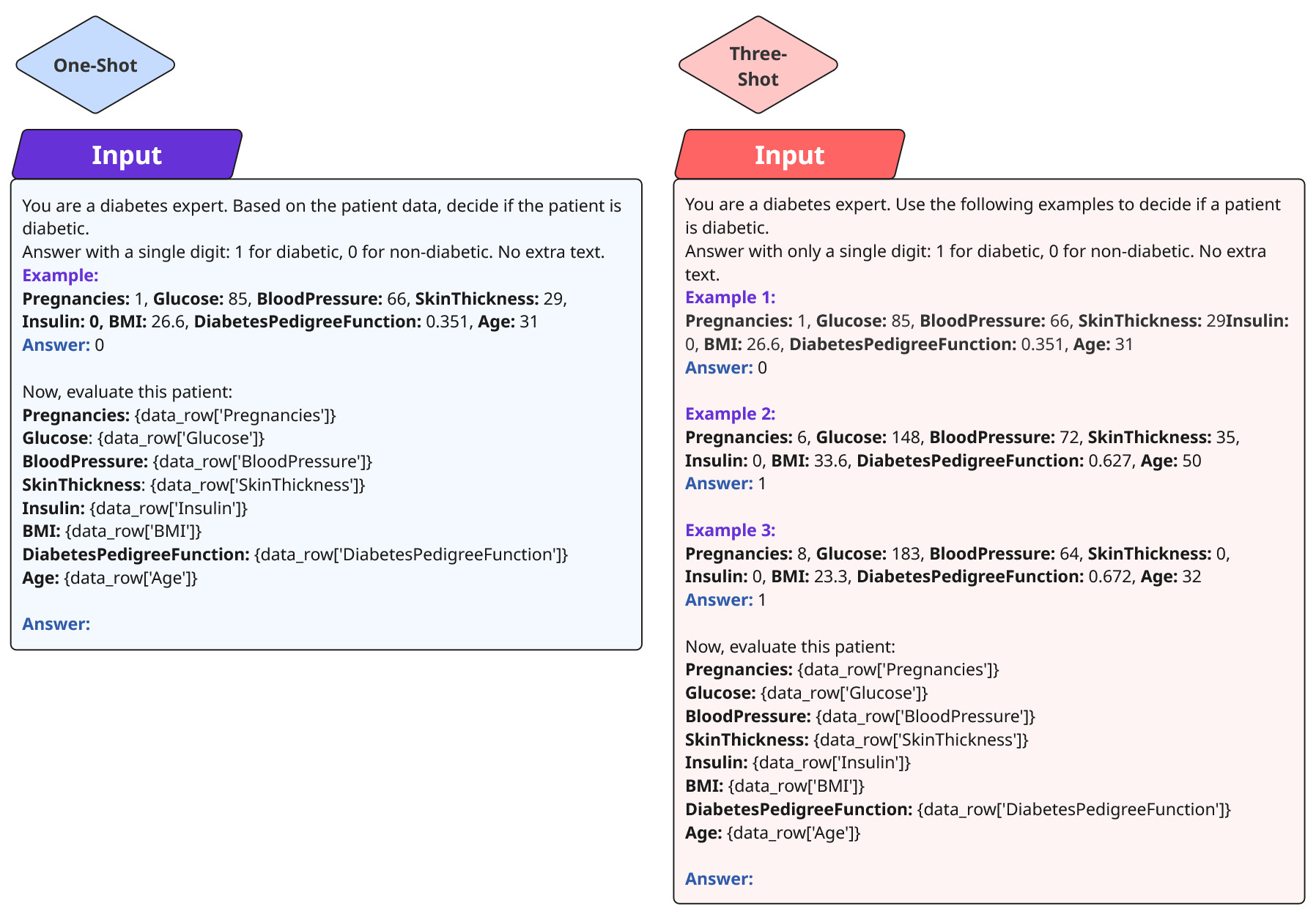}
  \caption{One-shot and Three-shot prompts in a diabetes prediction.}
  \label{fig:research_methodology}
\end{figure*}

\subsubsection{Free and Open-Source Models}
\begin{itemize}
    \item \textbf{Gemma-2-27B:} Gemma-2-27B is a state-of-the-art open-source language model (2B–27B parameters) that employs advanced attention mechanisms to deliver performance comparable to models two to three times larger, excelling in specialized domains \cite{gemma2024}.
    \item \textbf{Mistral-7B:} Mistral-7B is an open-source language model with 7 billion parameters, built on a dense Transformer architecture to ensure efficient and robust performance across a variety of NLP tasks \cite{mistral7b}.
    \item \textbf{Llama 3.1 8B:} Llama 3.1 8B is a variant from the Llama series featuring 8 billion parameters, designed for balanced performance and efficiency using advanced transformer architectures \cite{llama31}.
    \item \textbf{Llama 3.2 2B:} Llama 3.2 2B is a compact variant from the Llama family, with 2 billion parameters optimized for resource-constrained environments, yet still delivering competitive performance \cite{llama32}.
\end{itemize}

\subsubsection{Proprietary Models}

\begin{itemize}
    \item \textbf{GPT 4o:} GPT 4o is an advanced proprietary language model from the OpenAI GPT family, built on a large-scale transformer architecture. It excels in generative and reasoning tasks, offering significant improvements in contextual understanding and safety \cite{gpt4o}.
    \item \textbf{Gemini Flash 2.0:} Gemini Flash 2.0 is an advanced multimodal language model developed by Google DeepMind, leveraging flash attention mechanisms for efficient inference. It delivers rapid, robust performance with high accuracy and state-of-the-art capabilities in complex tasks \cite{geminiflash2}.
\end{itemize}

These open-source models were accessed and deployed using the Ollama platform within Kaggle's cloud-based environment. Ollama facilitates running LLMs on cloud platforms like Kaggle, providing a REST API for interaction. This setup allows for flexible experimentation and integration into various applications without the need for local hardware resources.\footnote{\url{https://ollama.com}} For the proprietary models, GPT-4o and Gemini Flash 2.0, we utilized their respective APIs to perform inference tasks.

Each LLM and prompting configuration (zero-shot, one-shot, three-shot) was executed three times using independently shuffled test sets to account for variability. The final reported metrics reflect the average performance across these runs.

\subsection{Prompt Design and Evaluation}

To assess the adaptability of LLMs to structured numerical data, we designed three distinct prompting strategies: zero-shot, one-shot, and three-shot, each reflecting a different level of context and task instruction. Since LLMs are inherently text-based and not trained on tabular data representations, we converted each data sample into a semantically meaningful natural language prompt using plain-text formatting.

\begin{itemize}
    \item \textbf{Zero-shot Prompting:}  
    The model is given a single, unlabeled patient input expressed in natural language. The prompt includes an instruction followed by the patient features, as shown in the example box below. No prior examples or additional context are provided. This setup allows us to assess the model’s inherent knowledge and ability to generalize from a direct prompt.

    \begin{tcolorbox}\footnotesize\sffamily
    \texttt{Predict whether the patient has diabetes based on the following information: Pregnancies: 3, Glucose: 120, BloodPressure: 70, SkinThickness: 25, Insulin: 80, BMI: 30.5, DiabetesPedigreeFunction: 0.5, Age: 33. Return 0 for non-diabetic and 1 for diabetic.}
    \end{tcolorbox}

    \item \textbf{One-shot Prompting:}  
    Before processing new patient data, the model is given a single labeled example along with a task-specific prompt. This minimal demonstration helps the model understand the expected format and improves its ability to apply relevant knowledge while still relying primarily on its pre-trained understanding.

    \item \textbf{Three-shot Prompting:}  
    The model is provided with three labeled examples from the dataset, along with a structured prompt that reinforces the task definition. This additional context enhances the model’s ability to recognize patterns, refine its predictions, and improve adaptability to the classification task.
\end{itemize}

In each case, the features were written in full, using plain descriptors and numeric values, to ensure maximum interpretability by the language model. All LLMs were instructed to return only a binary value (0 or 1) to facilitate clean evaluation against the ground truth labels. Figure~\ref{fig:research_methodology} illustrates the structure of One-shot and Three-shot prompts used in our evaluation pipeline.

Although no post-processing was applied to enforce binary formatting, all LLM responses conformed to the expected 0/1 outputs. This consistency is likely due to the structured nature of the input data and the constrained prompt design. Prior work has demonstrated that LLMs exhibit more reliable and deterministic behavior in low-entropy, numerically grounded tasks compared to open-ended natural language generation \cite{wang2023llmstructured}.

\subsection{Evaluation Metrics}  

To assess the performance of LLMs in diabetes prediction, we use four standard evaluation metrics: accuracy, precision, recall, and F1-score \cite{rainio2024evaluation}. These metrics provide a comprehensive analysis of classification effectiveness, considering both correct classifications and potential misclassifications.  

\textbf{Accuracy}: Measures the proportion of correctly classified instances among all instances. It is defined as:  
\begin{equation}  
\text{Accuracy} = \frac{TP + TN}{TP + TN + FP + FN}  
\end{equation}  
where $TP$ and $TN$ represent the correctly predicted diabetic and non-diabetic cases, while $FP$ and $FN$ denote false positives and false negatives, respectively.  

\textbf{Precision}: Represents the proportion of true positive predictions among all positive predictions, highlighting the model’s ability to avoid false positives:  
\begin{equation}  
\text{Precision} = \frac{TP}{TP + FP}  
\end{equation}  

\textbf{Recall (Sensitivity)}: Measures the proportion of actual positive cases that are correctly identified by the model:  
\begin{equation}  
\text{Recall} = \frac{TP}{TP + FN}  
\end{equation}  

\textbf{F1-score}: Provides a balanced measure by combining precision and recall, especially useful in imbalanced datasets:  
\begin{equation}  
\text{F1-score} = \frac{2 \times \text{Precision} \times \text{Recall}}{\text{Precision} + \text{Recall}}  
\end{equation}  

These metrics ensure a robust evaluation of LLMs' effectiveness in structured numerical data classification, providing insights into their strengths and limitations in medical prediction tasks. For LLMs, reported scores are averages of three independent runs. For ML baselines, evaluation is based on a fixed 80/20 split with standard preprocessing.

\subsection{Evaluation of LLMs and Traditional ML Models}

To evaluate the effectiveness of LLMs for structured medical prediction, we assessed six LLMs across three prompting configurations—zero-shot, one-shot, and three-shot—on the PIDD. Each configuration was executed over three independent runs to mitigate variance due to randomness in model responses. We report average values for accuracy, precision, recall, and F1-score.

In parallel, we implemented three traditional ML models—Logistic Regression, Random Forest, and SVM—as baselines. These were trained using an 80/20 train-test split with standard z-score normalization. This dual-track evaluation allows us to assess the strengths of LLMs in zero-training scenarios against well-established trained classifiers.

\autoref{table:combined_performance} presents the consolidated performance of all models. Across nearly all LLMs, three-shot prompting consistently led to performance improvements, validating the importance of in-context learning \cite{lakera2024incontext}. Among LLMs, Gemma-2-27B (three-shot) achieved the highest accuracy at 74.35\%, followed closely by GPT-4o (three-shot) at 74.22\%. Notably, both LLMs outperformed traditional models in F1-score, suggesting better balance between precision and recall.

To aid interpretation, the best-performing scores for each metric have been visually highlighted within the table.

\FloatBarrier
Despite slightly lower accuracy than Random Forest (75.97\%), LLMs demonstrated strong potential without any model-specific training.

\begin{table*}[!t]
\centering
\caption{Performance of LLMs and Traditional ML Models on the PIDD Dataset}
\label{table:combined_performance}
\begin{adjustbox}{max width=\linewidth}
\renewcommand{\arraystretch}{1.2}
\setlength{\tabcolsep}{5pt} 
\begin{tabular}{l l c c c c}
\toprule
\textbf{Category} & \textbf{Model} & \textbf{Accuracy} & \textbf{Precision} & \textbf{Recall} & \textbf{F1-Score} \\
\midrule
\multirow{18}{*}{Large Language Models} 
& Gemma-2-27B (Zero-shot)        & 0.7201 & 0.7159 & \cellcolor{yellow!20}\textbf{0.7374} & 0.7122 \\
& Gemma-2-27B (One-shot)         & 0.6549 & \cellcolor{yellow!20}\textbf{0.7232} & 0.7220 & 0.6549 \\
& {Gemma-2-27B (Three-shot)}       & 0.7435 & 0.7212 & 0.7320 & \cellcolor{yellow!20}\textbf{0.7250} \\
\cmidrule(lr){2-6}
& GPT-4o (Zero-shot)             & 0.7135 & 0.6938 & 0.7073 & 0.6969 \\
& GPT-4o (One-shot)              & 0.7227 & 0.6986 & 0.7074 & 0.7018 \\
& GPT-4o (Three-shot)            & 0.7422 & 0.7197 & 0.7301 & 0.7234 \\
\cmidrule(lr){2-6}
& Mistral-7B (Zero-shot)         & 0.4193 & 0.6786 & 0.5531 & 0.3714 \\
& Mistral-7B (One-shot)          & 0.3893 & 0.6818 & 0.5310 & 0.3250 \\
& Mistral-7B (Three-shot)        & 0.6016 & 0.6824 & 0.6732 & 0.6011 \\
\cmidrule(lr){2-6}
& Llama-3.1-8B (Zero-shot)       & 0.3646 & 0.6144 & 0.5103 & 0.2879 \\
& Llama-3.1-8B (One-shot)        & 0.3516 & 0.6749 & 0.5020 & 0.2632 \\
& Llama-3.1-8B (Three-shot)      & 0.3568 & 0.6759 & 0.5060 & 0.2721 \\
\cmidrule(lr){2-6}
& Llama-3.2-2B (Zero-shot)       & 0.3503 & 0.6747 & 0.5010 & 0.2609 \\
& Llama-3.2-2B (One-shot)        & 0.3542 & 0.6754 & 0.5040 & 0.2676 \\
& Llama-3.2-2B (Three-shot)      & 0.3698 & 0.6094 & 0.5134 & 0.2977 \\
\cmidrule(lr){2-6}
& Gemini Flash 2.0 (Zero-shot)   & 0.7331 & 0.4710 & 0.4682 & 0.4695 \\
& Gemini Flash 2.0 (One-shot)    & 0.7292 & 0.4684 & 0.4674 & 0.4679 \\
& Gemini Flash 2.0 (Three-shot)  & 0.7305 & 0.4703 & 0.4727 & 0.4713 \\
\midrule
\multirow{3}{*}{Machine Learning Algorithms} 
& {Random Forest}                  & \cellcolor{yellow!20}\textbf{0.7597} & 0.6552 & 0.6909 & 0.6726 \\
\cmidrule(lr){2-6}
& Logistic Regression            & 0.7532 & 0.6491 & 0.6727 & 0.6607 \\
\cmidrule(lr){2-6}
& SVM                            & 0.7338 & 0.6458 & 0.5636 & 0.6019 \\
\bottomrule
\end{tabular}
\end{adjustbox}
\end{table*}
\FloatBarrier

\begin{figure}[H]
\centering
\includegraphics[width=\linewidth]{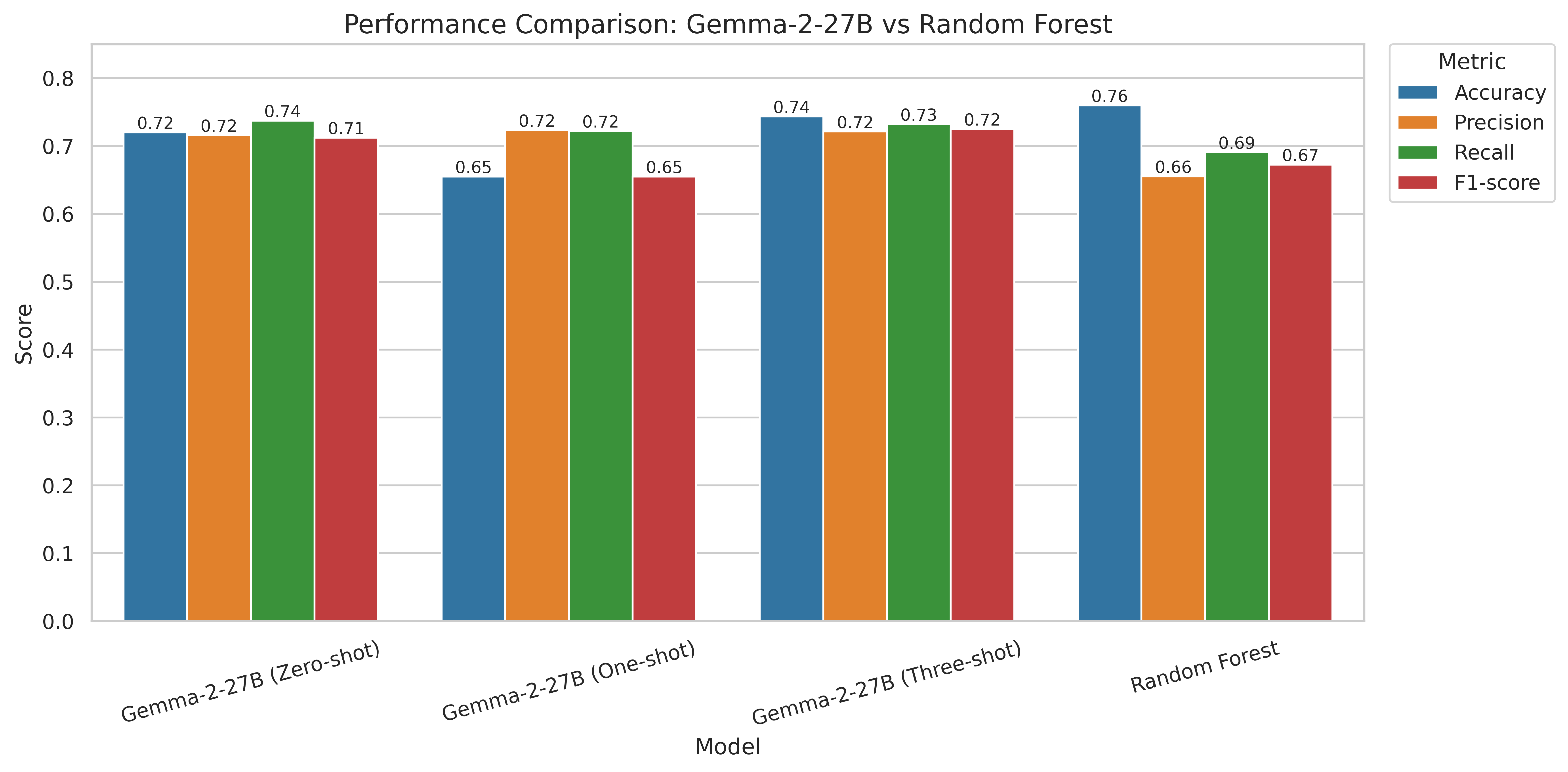}
\caption{Comparison of performance metrics between Gemma-2-27B (across prompt types) and Random Forest on the PIDD dataset.}
\label{fig:gemma_vs_rf}
\end{figure}

To complement the tabular results, \autoref{fig:gemma_vs_rf} provides a visual comparison of the performance metrics achieved by the Gemma-2-27B model under varying prompt configurations (zero-shot, one-shot, and three-shot) against the highest-performing traditional baseline, Random Forest. The bar chart offers an intuitive perspective on the relative strengths of each approach across accuracy, precision, recall, and F1-score.

\vspace{-4pt}
\vspace{-4pt}

\subsection{Observations and Implications}

Prompt engineering significantly affects LLM performance in structured tasks like medical classification. GPT-4o and Gemma-2-27B show strong potential in low-data scenarios, while Llama and Mistral underperform without task-specific tuning.

LLMs operate without explicit training but demand more resources than models like Random Forest or SVM. Gemini Flash 2.0’s high accuracy but low F1-score reflects a tendency to overpredict the majority class, influenced by the dataset’s class imbalance.

As seen in Table~\ref{table:combined_performance}, improvements from three-shot prompting were not consistent across models, suggesting that one or three examples are not sufficient to yield reliable gains without more context or tuning.

\subsection{Comparison with Existing Literature}

In contrast to earlier studies reporting up to 98.07\% accuracy with deep learning \cite{naz2020deep} and 90.7\% with hybrid RFE-GRU models \cite{shams2025novel}, the LLMs in our evaluation achieved a maximum accuracy of 74.35\%. Although these results do not surpass those of specialized ML/DL systems, they suggest that LLMs can offer reasonable performance without requiring labeled training data or task-specific model tuning. This indicates their potential as flexible tools for initial experimentation, especially in resource-constrained or data-scarce scenarios.

\subsection{Future Works}

Although prompt-based LLMs showed encouraging results, their performance remains below that of traditional machine learning methods. Future research should explore fine-tuning LLMs on structured datasets and investigate hybrid approaches that combine LLMs with conventional classifiers. These directions may help bridge the performance gap and enhance applicability in medical prediction tasks.

\section{Conclusion}

This study evaluated prompt-based LLMs for diabetes prediction using the PIDD dataset, comparing their performance to traditional machine learning models like Logistic Regression, Random Forest, and SVM. Our findings show that while LLMs, particularly Gemma-2-27B and GPT-4o, perform competitively in few-shot scenarios, they still fall short of surpassing classical models in accuracy. The ability of LLMs to work without training and respond to structured prompts presents advantages in exploratory or resource-limited environments. Additionally, three-shot prompting consistently enhanced model performance. However, challenges such as high computational cost, prompt sensitivity, and limited interpretability highlight that prompt-based LLMs are not yet viable replacements for conventional methods in structured medical predictions.

\bibliography{Ref}
\end{document}